\documentclass[letterpaper,10pt,twocolumn,oneside,final,conference]{ieeeconf}  

\IEEEoverridecommandlockouts                              

\usepackage{graphicx}
\usepackage{amsmath}
\usepackage{amssymb}
\usepackage{subcaption}

\title{\LARGE \bf
Discovering Predictive Relational Object Symbols\\with Symbolic Attentive Layers
}

\author{Alper Ahmetoglu$^{1}$, Batuhan Celik$^{1}$, Erhan Oztop$^{2,3}$, Emre Ugur$^{1}$
\thanks{$^{1}$Department of Computer Engineering, Bogazici University
        {\tt\small alper.ahmetoglu@boun.edu.tr}}%
\thanks{$^{2}$Department of Computer Science, Ozyegin University}%
\thanks{$^{3}$OTRI, SISReC, Osaka University}
}

\begin{document}

\maketitle
\pagestyle{empty}

\begin{abstract}
In this paper, we propose and realize a new deep learning architecture for discovering symbolic representations for objects and their relations based on the self-supervised continuous interaction of a manipulator robot with multiple objects on a tabletop environment. The key feature of the model is that it can handle a changing number number of objects naturally and map the object-object relations into symbolic domain explicitly. In the model, we employ a self-attention layer that computes discrete attention weights from object features, which are treated as relational symbols between objects. These relational symbols are then used to aggregate the learned object symbols and predict the effects of executed actions on each object. The result is a pipeline that allows the formation of object symbols and relational symbols from a dataset of object features, actions, and effects in an end-to-end manner. We compare the performance of our proposed architecture with state-of-the-art symbol discovery methods in a simulated tabletop environment where the robot needs to discover symbols related to the relative positions of objects to predict the observed effect successfully. Our experiments show that the proposed architecture performs better than other baselines in effect prediction while forming not only object symbols but also relational symbols. Furthermore, we analyze the learned symbols and relational patterns between objects to learn about how the model interprets the environment. Our analysis shows that the learned symbols relate to the relative positions of objects, object types, and their horizontal alignment on the table, which reflect the regularities in the environment.
\end{abstract}

\section{INTRODUCTION}
\label{sec:intro}
Learning the symbolic representation of tasks enables the application of classical AI search techniques to find a solution in the symbolic definition of the task. For well-defined environments, symbolic systems can be manually designed to describe robot-environment interactions. However, such manual designs would only be scalable to a handful of domains and require significant work to adapt to new environments. On the other hand, learning the required symbols for the task from data would be a more scalable and generalizable strategy to achieve truly intelligent robots \cite{konidaris2019necessity}. Therefore, there is a considerable amount of research on how to convert the sensorimotor experience of a robotic agent into symbolic representations \cite{taniguchi2018symbol}.

One prominent strategy for learning the necessary symbols is to partition the precondition and the effect set of the agent's actions and learn classifiers for these partitions
\cite{konidaris2014constructing,konidaris2018skills,Ugur-2015-ICRA}. This ensures that the learned symbols are compatible with the actions available to the agent and filters out irrelevant aspects of the environment that the agent cannot manipulate. Learning symbols can also be formalized as compressing the state-space into symbolic state-space with autoencoders \cite{asai2018classical,asai2020learning,asai2022classical}. Operators, which are high-level actions manipulating these symbols, can be learned simultaneously or separately. One of the main advantages of this approach is that symbols are learned with deep neural networks, which opens up the possibility of integrating the recent advances in deep learning.

\begin{figure*}[htbp]
    \centering
    \includegraphics[width=0.99\linewidth]{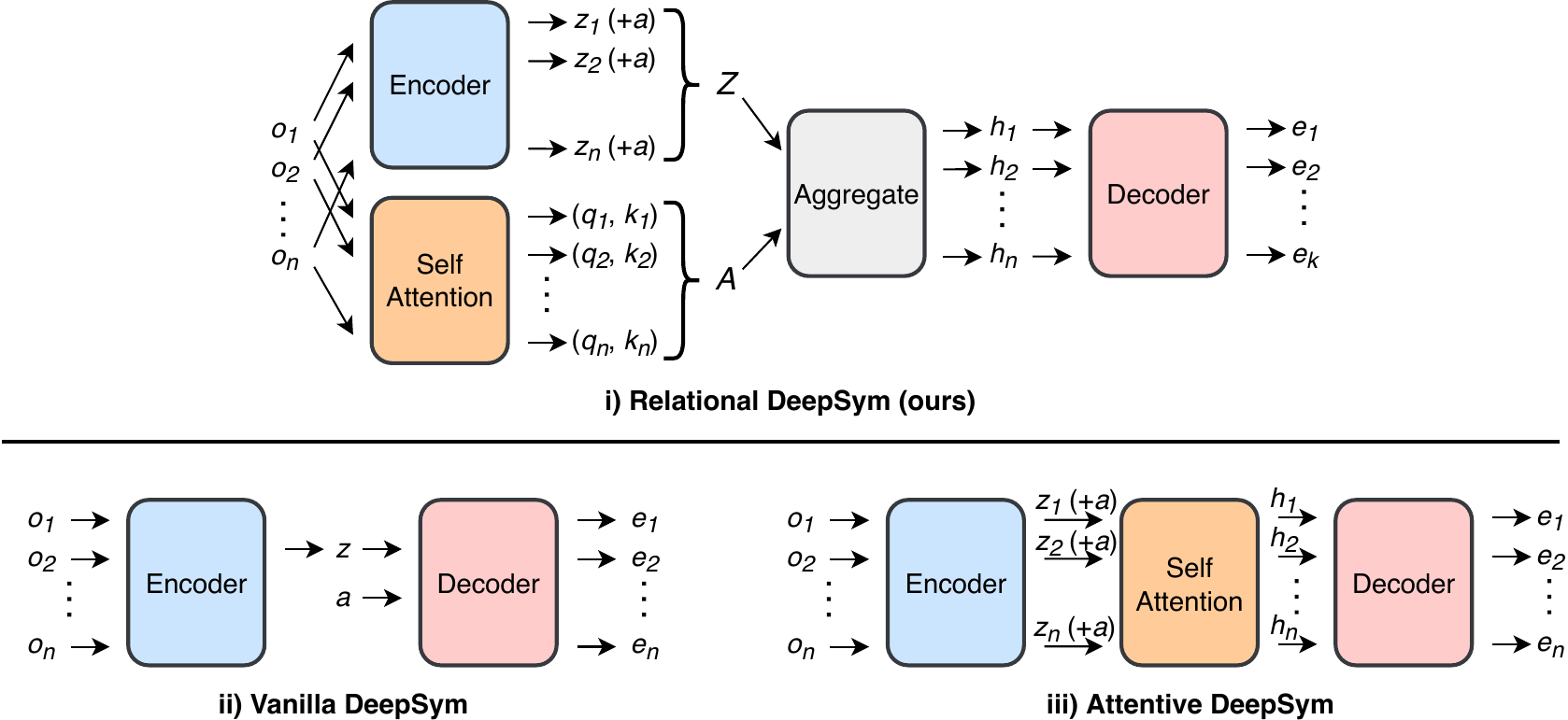}
    \caption{The proposed model is shown in the top panel. The encoder and the self-attention module take object features as input and process them in parallel. The encoder outputs an object symbol $z_i$ for the object $o_i$, and the self-attention module outputs the query vector $q_i$ and the key vector $k_i$ which are used as in Equation \ref{eq:qk} to calculate relational symbols. For comparison, we also provide high-level outlines of \cite{ahmetoglu2022deepsym} and \cite{ahmetoglu2022learning} in the bottom panel in (ii) and (iii), respectively.}
    \label{fig:model}
\end{figure*}

Our previous work, DeepSym \cite{ahmetoglu2022deepsym}, combines these two motivations: learning preconditions and effects of actions with deep neural networks. In DeepSym, an encoder-decoder network with a discrete bottleneck layer is trained to predict the effect of actions (Figure \ref{fig:model} -- bottom left). However, the network can only handle a fixed number of object interactions, restricting the types of relations that can be learned. This restriction is lifted in a more recent work \cite{ahmetoglu2022learning} by introducing a self-attention mechanism \cite{vaswani2017attention} to the architecture (Figure \ref{fig:model} -- bottom right).  As symbols interact with each other using self-attention, the network can make accurate predictions for related objects (e.g., on top of each other). Although this architecture is effective in making accurate predictions for related objects through the learned multi-object symbols, it does do not reveal the explicit relations between objects. Furthermore, as the self-attention layer is applied after discretization, the  relational representational capacity of the model is limited by the learned symbols.

In the current work, we explicitly compute discrete self-attention weights from object features and treat them as relational symbols between objects. Using these discrete relations, we fuse object symbols in an aggregation function to produce a single representation for each object, which is then used to predict the observed, potentially multi-object, effect. This results in a more powerful architecture, which we named Relational DeepSym, that can explicitly output the relations between a varying number of objects while enjoying other properties of DeepSym. Our experiments in a simulated tabletop scenario show that (1) Relational DeepSym achieves lower errors than \cite{ahmetoglu2022deepsym} and \cite{ahmetoglu2022learning} for different numbers of objects and actions, (2) learns not only object symbols but also relational symbols, (3) and these learned object and relational symbols contain information about object types and their relative placements that are in line with the given task.

\section{RELATED WORK}
\label{sec:related}

Early symbol grounding studies in robotics (e.g., \cite{mourao2008,worgotter2009}) assumed the existence of manually defined symbols that were effective in plan generation. These studies collected data from interactions of agents and robots and learned sensor-to-symbol mappings to ground the pre-defined symbols in the sensorimotor experience of the robot. In these studies, transition rules, which are connected by symbolic preconditions and effects, were defined, and the continuous experience of the robot was used to map the manually defined symbolic predicates to the continuous perceptual space of the robots. Recently, \cite{dehban2022} proposed a deep neural network architecture based on Convolutional Variational Auto-Encoders to discover visual features that are well-suit for pre-defined recognition and interaction tasks. \cite{lay2022unsupervised} used Multi-modal Latent Dirichlet Allocation (MLDA)  to learn the mapping between multi-modal sensory experience and preconditions and post-conditions of actions of a robot. We argue that pre-defining symbols in unknown and changing environments is not possible, and as stated by \cite{sun2000symbol}, symbols should instead ``be formed in relation to the experience of agents, through their perceptual/motor apparatuses, in their world and linked to their goals and actions''.

Unsupervised discovery of discrete symbols and rule learning from the continuous sensorimotor experience of embodied agents has been recently studied in robotics in order to equip robots with advanced reasoning and planning capabilities \cite{taniguchi2018symbol,konidaris2019necessity}. \cite{ahmetoglu2022high} investigated the discovery of sub-symbolic neural activations that facilitate resource economy and fast learning in skill transfer but did not address high-level reasoning with discrete symbols. \cite{konidaris2014constructing,Konidaris2015} discovered symbols that were directly used as predicates in precondition and post-condition fields of action descriptors, represented in Problem Domain Definition Language (PDDL). This encoding allowed for making deterministic and probabilistic plans in 2-dimensional agent environments. The same architecture was extended to a real-world robotic environment in \cite{konidaris2018skills}, where symbols representing absolute global states were learned and used for planning. \cite{james2020learning}, on the other hand, learned egocentric symbolic representations that enabled the agents to transfer the previously learned symbols to novel environments directly. \cite{james2021autonomous} considers learning symbols from object-centric observations, allowing for a transfer between tasks that share the same types of objects. Effect clustering techniques and SVM classifiers were used to discretize the continuous sensorimotor experience of the agents in these works.

\begin{figure*}[htbp]
    \centering
    \includegraphics[width=0.99\linewidth]{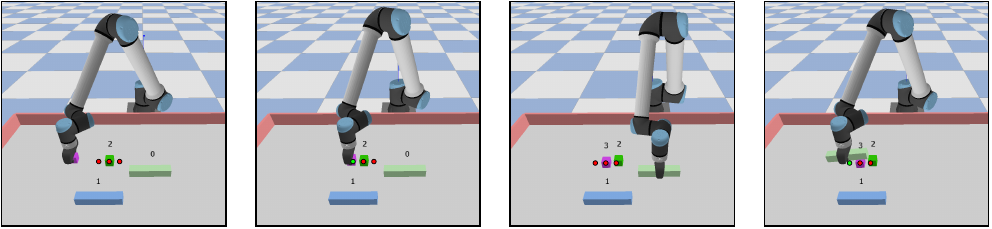}
    \caption{An example of the data collection process. The red dots show possible release positions, and the green dot shows the selected release position.}
    \label{fig:setup}
\end{figure*}

Whereas the previous work addressed learning symbols from given skills, \cite{silver2021learning,chitnis2022learning,silver2022learning,li2023embodied,achterhold2023learning} learned a set of skills from a set of symbolic predicates and a collection of demonstrations. \cite{kumar2023overcoming} considered only the necessary changes in the predicate set for more compact operators. In follow-up work, \cite{silver2023predicate} used a surrogate objective for learning state abstractions that increase planning performance. \cite{Ugur-2015-ICRA,Ugur-2015-Humanoids} discovered discrete symbols and used these symbols in order to generate PDDL rules for planning by again combining effect clustering techniques to find discrete effect categories and SVM classifiers to discretize continuous object feature space. These studies used ad-hoc combinations of several machine learning methods. On the other hand, \cite{ahmetoglu2022deepsym} provided a more generic symbol formation engine, which used a novel deep network architecture that runs at the pixel level and relies on purely predictive mechanisms in forming symbols instead of unsupervised clustering techniques. They used an effect predictor encoder-decoder network that took the object image and action as input and exploited a binary bottleneck layer to automatically form object categories. Similar to this work, \cite{asai2018classical,asai2020learning,asai2022classical} also exploited deep neural networks with binary bottleneck units to find discrete state and effect symbols and achieve plan generation using these symbols. \cite{tekden2020belief,tekden2021object} model multi-object dynamics using graph neural networks and predict object relations with a recurrent architecture. Similarly, \cite{huang2023planning} learns a latent state-space in a graph structure and a transition function that can predict object relations between objects after an action execution given the graph.

Our work differs from previous research as we propose a new method for learning symbolic representations of objects and relations between them in a unified architecture. Most related to \cite{ahmetoglu2022learning} that also uses self-attention to model relational information, our model differs in that it explicitly outputs the relations between objects. In contrast, in \cite{ahmetoglu2022learning}, the learned relations are opaque to the user.

\section{METHOD}
\label{sec:method}

The problem definition and our assumptions are given in Section \ref{subsec:problem}, the proposed model is explained in Section \ref{subsec:method}, and the differences with previous DeepSym architectures are discussed in Section \ref{subsec:comparison}.
\subsection{Problem Definition}
\label{subsec:problem}
This work deals with the problem of learning symbolic representations of objects and relations between them from continuous state representations collected by a robot to predict the effect of its actions. From a developmental learning perspective, this study starts off with a basic sensorimotor system \cite{ugur2015staged,ugur2015parental}, where 
the robot can locate objects, and pick and place them on top of each other. We compare our proposed architecture with \cite{ahmetoglu2022deepsym} and \cite{ahmetoglu2022learning} in a tabletop environment by their effect prediction performance.

\subsection{Relational DeepSym}
\label{subsec:method}
The top panel in Figure \ref{fig:model} shows a high-level overview of the proposed model. We define the state vector as a set of object poses and types $\{o_1, o_2, \dots, o_n\}$ where $n$ is the number of objects that varies in each sample. The model consists of four main components: (1) an encoder that transforms the state vector into symbolic representations $\{z_1, z_2, \dots z_n\}$ that are fixed-sized binary vectors for each object, (2) a self-attention module that outputs query and key vectors for each object from the state vector (3) an aggregation function that combines information from multiple objects by multiplying object symbols with relational symbols, and (4) a decoder that predicts the generated effect of the executed action for each object. As the whole architecture is differentiable and trained in an end-to-end fashion to minimize the effect prediction error, we expect the encoder and the self-attention module to learn to predict symbols and relations useful for the decoder to predict the effect.

The encoder takes the state vector $\{o_1, o_2, \dots, o_n\}$ as input and processes them independently to output a set of discrete vectors $\{z_1, z_2, \dots, z_n\}$ that are treated as object symbols. To output a discrete vector without removing the differentiability, the activation of the last layer is set to the Gumbel-sigmoid function \cite{maddison2016concrete,jang2016categorical}. In our experiments, the encoder is a multi-layer perceptron; however, other differentiable architectures can be used for different modalities (e.g., convolutional layers to process images)

\begin{figure*}[htbp]
    \centering
    \includegraphics[width=0.99\textwidth]{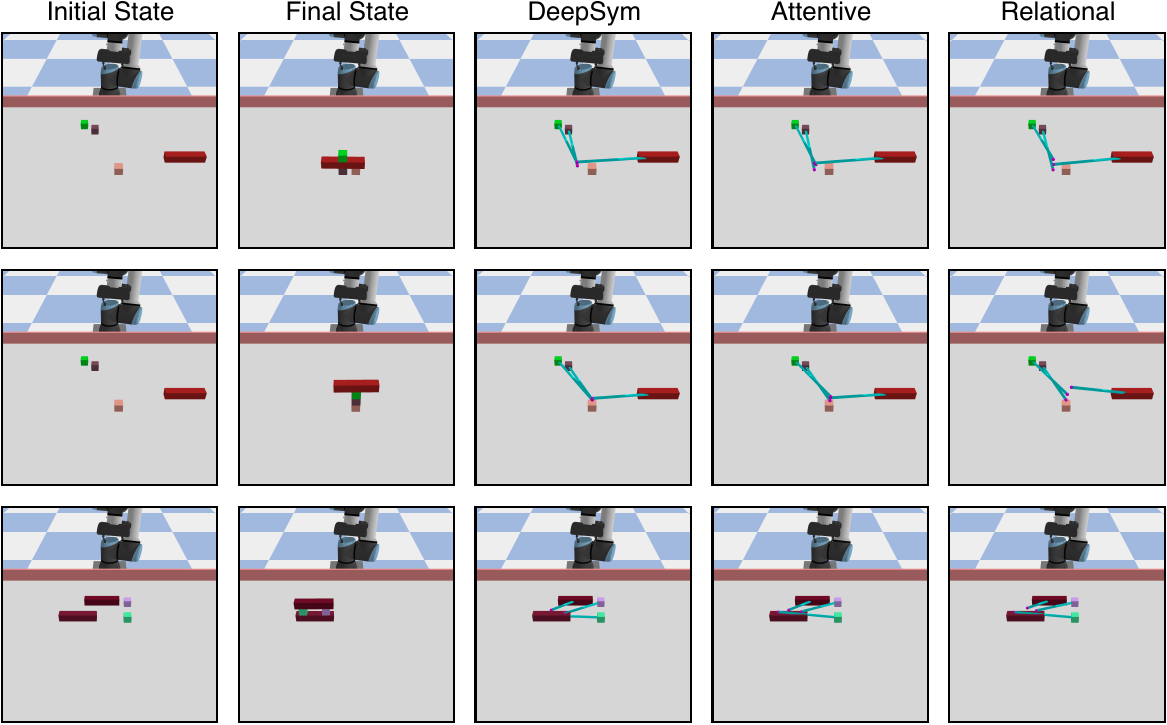}
    \caption{Action sequence prediction results for different models. The initial state (the input to models) is shown in the first column. The second column shows the ground truth final state after executing a sequence of actions. Columns 3-5 show the predicted object movements for different models.}
    \label{fig:preds}
\end{figure*}

The self-attention module takes object features $\{o_1, o_2, \dots, o_n\}$ as input and processes them independently to output query and key vectors $\{(q_1, k_1), (q_2, k_2), \dots, (q_n, k_n)\}$ for each object. Let $Q$ and $K$ be $n \times d$ matrices, each row containing a query vector $q_i$ and a key vector $k_i$, respectively. The attention weights $A$ are computed as follows:
\begin{equation}
    A = \text{GumbelSigmoid}\left(\frac{QK^T}{\sqrt{d}}\right)
    \label{eq:qk}
\end{equation}
where $d$ is the dimensionality of the query and key vectors. This is slightly different from the regular self-attention function \cite{vaswani2017attention} in which the softmax function is used instead of the Gumbel-sigmoid function. This modification creates two different behaviors: (1) the use of a sigmoid function instead of a softmax function allows multiple attention weights to be active at the same time (whereas in softmax, attentions compete with each other), and (2) the use of the Gumbel-sigmoid function discretizes the attention weights while preserving differentiability, allowing us to treat the weights as \emph{relational symbols} between objects. As in the encoder, the self-attention module is a multi-layer perceptron with two different outputs for the query and the key. Note that multiple heads $A_1, A_2, \dots, A_k$ can be used to model different relations between objects.

In the third step, the aggregation function combines object symbols $\{z_1, z_2, \dots, z_n\}$, relational symbols $\{A_1, A_2, \dots A_k\}$, and the executed action $a$ to produce a single representation for each object. The aggregation function has the following steps:
\begin{align}
    \bar{z}_i &= \text{concat}(z_i, a)\label{eq:concat} &\forall i \in \{1, 2, \dots, n\}\\
    h^{j}_i &= \text{matmul}(A_j, \text{mlp}(\bar{z}_i))\label{eq:agg} &\forall i \in \{1, 2, \dots, n\}\notag\\
    &&\forall j \in \{1, 2, \dots, k\}\\
    h_i &= \text{concat}(h^1_i, h^2_i, \dots, h^k_i) & \forall i \in \{1, 2, \dots, n\}\label{eq:concat2}
\end{align}
where object symbols and the action vector are concatenated in Equation \ref{eq:concat}, and the aggregation occurs in Equation \ref{eq:agg}. Here, the action vector $a$ is assumed to be discrete. One can possibly aggregate the input multiple times by applying Equation \ref{eq:agg} more than once to model longer effect chains. In our experiments, we use a single aggregation step. Multiple combinations from multiple attention heads are concatenated in Equation \ref{eq:concat2} to produce a single representation $h_i$ for each object.

As the final step, the decoder takes the aggregated representation $h_i$ as input and predicts the effect $\hat{e}$ for each object for the executed action $a$. The decoder is a multi-layer perceptron. The predicted effect is then compared with the ground truth effect $e$ to compute the mean squared error:
\begin{equation}
    \mathcal{L} = \frac{1}{M}\sum_{j=1}^M \sum_{i=1}^N (\hat{e}^{(j)}_i - e^{(j)}_i)^2
    \label{eq:loss}
\end{equation}
where $M$ is the batch size, and $N$ is the number of objects.

\subsection{Comparison with Related Models}
\label{subsec:comparison}
As in DeepSym (Figure \ref{fig:model} -- bottom left), this architecture is also an encoder-decoder architecture with discrete bottleneck layers. The difference is that the information between objects is shared in the aggregation function using the learned attention weights for a more accurate effect prediction for actions involving several objects. In DeepSym, this can only be achieved by fixing the number of input objects, whereas there is no such limitation in the proposed model.

\begin{figure*}[htbp]
    \centering
    \begin{subfigure}[b]{0.24\textwidth}
        \centering
        \includegraphics[width=\textwidth]{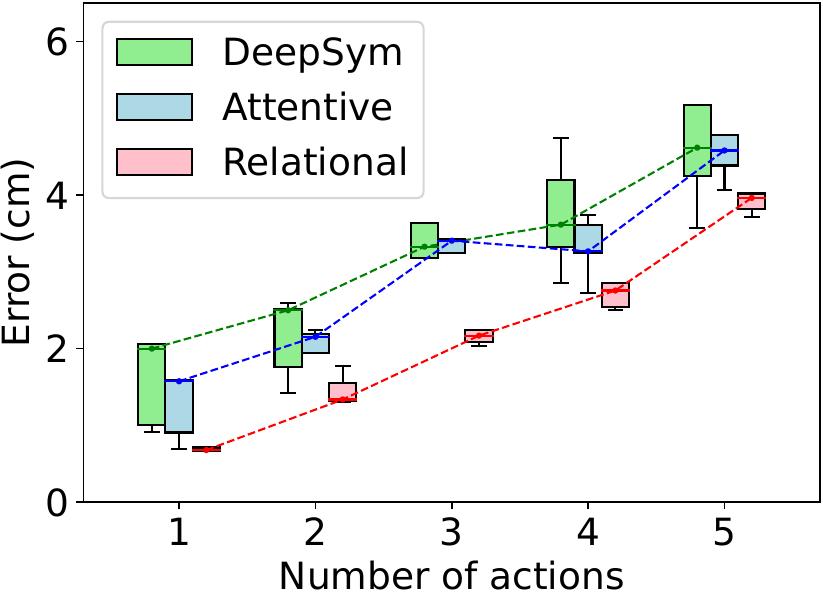}
        \caption{2 objects}
        \label{subfig:cum2obj}
    \end{subfigure}
    \hfill
    \begin{subfigure}[b]{0.24\textwidth}
        \centering
        \includegraphics[width=\textwidth]{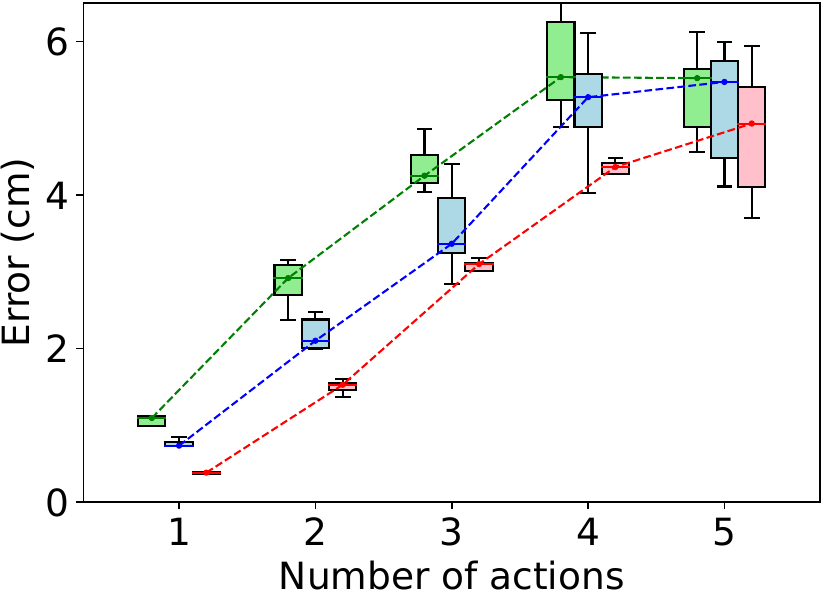}
        \caption{3 objects}
        \label{subfig:cum3obj}
    \end{subfigure}
    \hfill
    \begin{subfigure}[b]{0.24\textwidth}
        \centering
        \includegraphics[width=\textwidth]{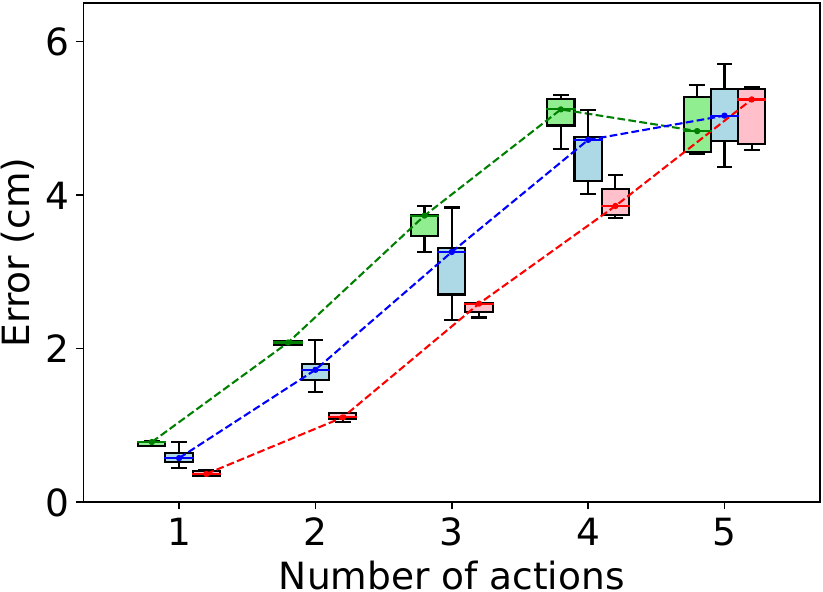}
        \caption{4 objects}
        \label{subfig:cum4obj}
    \end{subfigure}
    \hfill
    \begin{subfigure}[b]{0.24\textwidth}
        \centering
        \includegraphics[width=\textwidth]{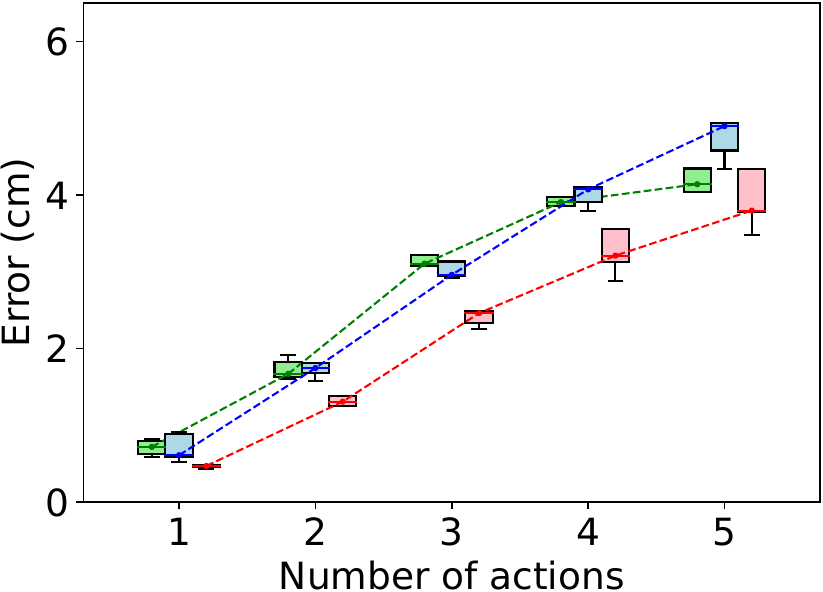}
        \caption{2-4 objects}
        \label{subfig:cum234obj}
    \end{subfigure}
    \caption{Prediction errors for different models as the number of actions increases.}
    \label{fig:errors}
\end{figure*}

Regarding the architecture in \cite{ahmetoglu2022learning} (Figure \ref{fig:model} -- bottom right), the most significant difference is the placement of the self-attention module. In \cite{ahmetoglu2022learning}, the self-attention module takes object symbols (the encoder's output) as its input and directly outputs the aggregated representation. This restricts the model from learning attention weights only from the learned symbols. In this proposal, attention weights are learned from object features, making relations more general.

The second significant difference is the explicit use of attention weights. In \cite{ahmetoglu2022learning}, attention weights are used within the self-attention layer as in the original Transformer architecture \cite{vaswani2017attention}. However, since attention weights are continuous, they cannot be easily expressed as relational symbols between objects.

\section{EXPERIMENTS}
\label{sec:exps}

\subsection{Experiment Setup}
\label{subsec:expsetup}

\subsubsection{Environment}
We created a tabletop object manipulation environment for our experiments (Figure \ref{fig:setup}). The environment consists of a UR10 robot and two to four objects. These objects are either short blocks or long blocks. The robot has a single type of high-level action: grasping and releasing an object on top of or near another object. We assume that object positions can be recognized by a separate module, and the robot can track the cartesian position change of these objects. What is to be learned is the effect of the executed action on each object in different configurations.

In our experiments, we first collect a fixed-size dataset required for the training by interacting with the environment and then train the model. Note that this procedure can be turned into a buffer-based training where the model training and the data collection are done in parallel, similar to many reinforcement learning setups.

\subsubsection{Data Collection}
The robot picks a random object from its center or 7.5cm left/right of the center and places it on top of or 7.5cm left (or right) of another random object (Figure \ref{fig:setup}). Object features before the execution of the action are recorded as the state vector. Here, object features are object types and poses with respect to the object frame that is going to be picked, which allows models to generalize to different object positions. Effects are the concatenation of (1) the position change of objects before the pick-up action and (2) the position change of objects after the release action. Such an effect representation filters out the movement effect of the object from the source location to the target location. In this way, the effect representation models what happens `immediately after the pick-up' and `immediately after the release' actions. Object and effect representations might have been selected as raw images as in \cite{ahmetoglu2022deepsym,ahmetoglu2022learning}; however, we opt for a simpler setup to compare different architectures in a controlled environment.

To compare different architectures in different settings, we collected three datasets that contain exactly two, three, or four objects. We combine these datasets to create a fourth dataset that contains a varying number of objects. Each dataset contains $(\{o_1, o_2, \dots, o_n\}, a, \{e_1, e_2, \dots, e_n\})$ triplets where $n$ is the number of objects. We collect 120K samples for two objects, 180K for three objects, and 240K for four objects. We use 80\% of samples for training, 10\% for validation, and 10\% for testing.

\subsubsection{Baselines}
We compare our method with \cite{ahmetoglu2022deepsym} and \cite{ahmetoglu2022learning}. As the vanilla DeepSym architecture requires a fixed-size input and output, we modified it to make it suitable for our experiments. Namely, a maximum number of objects is determined for a given training session. Then, the input (and the output) vector is reshaped into $[o_{\text{grasped}}, o_{\text{released}}, o_{\text{rest}}]$ where $o_{\text{grasped}}$ and $o_{\text{released}}$ are the object features of the grasped and released objects, respectively, and $o_{\text{rest}}$ is the object features of the remaining objects.

\subsubsection{Training Details}
All architectures are trained with the same hyperparameters throughout the text unless mentioned otherwise. We train models for 4000 epochs with five repetitions with different seeds. Adam optimizer \cite{kingma2014adam} is used with a batch size of 128 and a learning rate of 0.0001. All network components (e.g., encoder, decoder) consist of two hidden layers with 128 hidden units. The number of attention heads for attentive models is set to four. We clip gradients by their norm to 10. Extended experimental details (training logs, layer gradients, and other training options) can be found at Weights \& Biases\footnote{\texttt{https://api.wandb.ai/links/alper/xvpcogu1}} \cite{wandb}.

\begin{figure*}[htbp]
    \centering
    \includegraphics[width=0.95\linewidth]{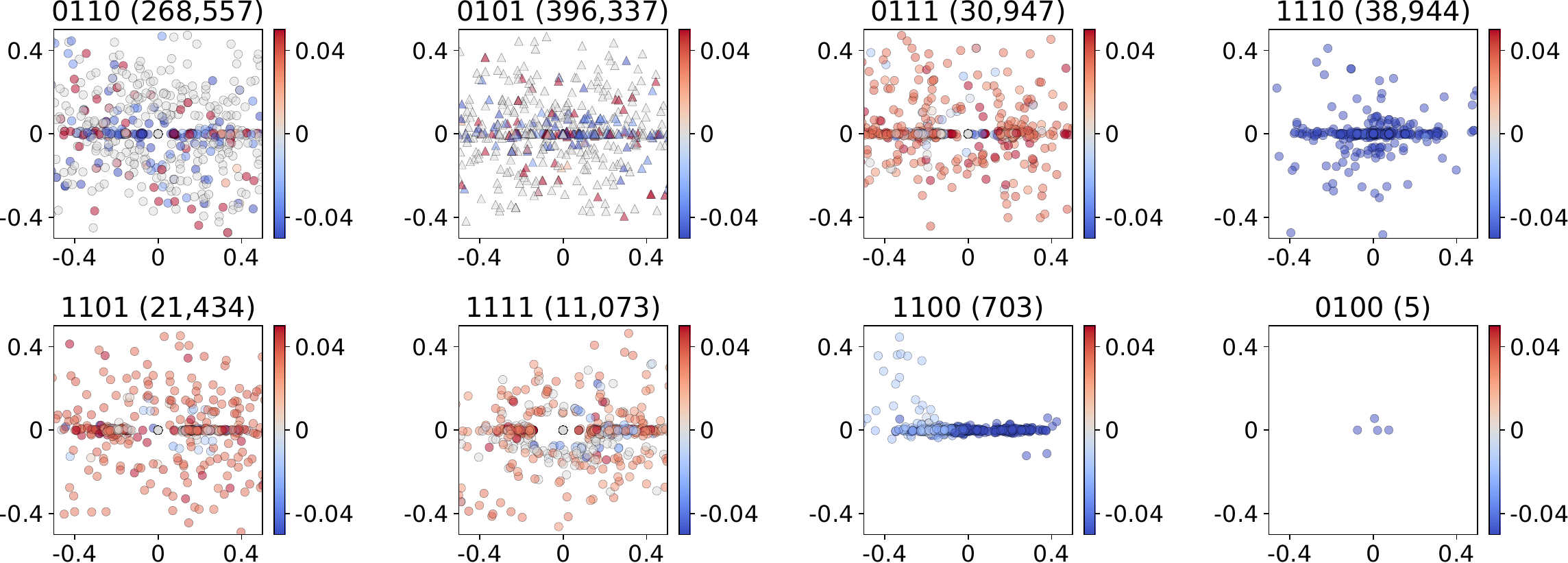}
    \caption{Each subfigure shows data points that activate the symbol given in the title together with the activation count. Circles and triangles represent short and long blocks, respectively. $x$ and $y$ axises represent the $x$ and $y$ positions of the objects while the color represents the $z$ position.}
    \label{fig:symbols}
\end{figure*}

\begin{figure*}[htbp]
    \centering
    \includegraphics[width=0.95\linewidth]{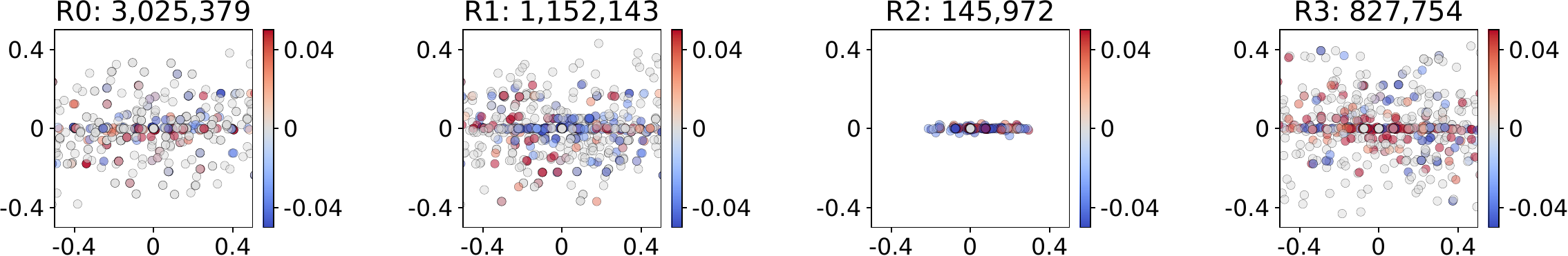}
    \caption{Each subfigure shows relative distances between objects that activate a relation. $x$ and $y$ axises represent the $\Delta x$ and $\Delta y$ between objects, and the color represents the $\Delta z$.}
    \label{fig:relations}
\end{figure*}

\subsection{Effect Prediction Results}
\label{subsec:results}
Firstly, we compare effect prediction results for different datasets in Table \ref{tab:results}. The reported results are absolute errors summed over all dimensions (three dimensions for the pick-up effect and three dimensions for the release effect). The results show that the proposed method achieves significantly lower errors than others. Moreover, the variance is lower than others, indicating that Relational DeepSym is more robust to different seeds.

\begin{table}[!t]
\caption{Effect prediction results averaged over five runs. Units are in centimeters. Welch's t-test \cite{welch1947generalization} shows significant differences ($p<0.02$ for all cases) between the proposed method and others.}
\begin{center}
\begin{tabular}{cccc}
\hline
Dataset & DeepSym & Attentive & Relational\\
\hline
2 objects & 2.22 $\pm$ 0.56 & 0.89 $\pm$ 0.10 & \textbf{0.50 $\pm$ 0.03}\\
3 objects & 3.06 $\pm$ 0.16 & 2.55 $\pm$ 0.09 & \textbf{1.67 $\pm$ 0.02}\\
4 objects & 4.26 $\pm$ 0.68 & 2.75 $\pm$ 0.12 & \textbf{2.00 $\pm$ 0.04}\\
2-4 objects & 2.38 $\pm$ 0.25 & 1.86 $\pm$ 0.12 & \textbf{1.35 $\pm$ 0.04}\\
\hline
\end{tabular}
\label{tab:results}
\end{center}
\end{table}

Errors increase as the number of objects increases. This is expected since the number of unique effects increases with the number of objects as the robot creates more complex structures in random exploration. Choosing the exploration schedule in a guided way, similar to experience replay in reinforcement learning \cite{schaul2015prioritized}, would be a promising future direction.

\subsection{Action Sequence Prediction}
\label{subsec:seqpred}
In this section, using effect predictions of models, we predict the next state by adding the prediction back into the state vector. Firstly, the predicted pick-up and release effects are summed with the state vector. Then, the movement from the pick-up position to the release position is added for objects that are predicted to be picked up. This way, given an initial state, we can predict the final state the robot reaches after executing a sequence of actions. Here, the challenge is to understand what happens when an object is lifted and released on top of another object in the presence of multiple objects.

Figure \ref{fig:preds} shows action sequence prediction examples. Relational DeepSym's predictions are more accurate than others, especially in the $z$-axis, the most significant axis in these experiments. This shows that the proposed model understands that the presence of an object on top of another object will change the action results.

In Figure \ref{fig:errors}, we analyze how models perform as the number of actions increases. We see that Relational DeepSym shows a slightly lower error than others. Errors increase for all models when the number of actions increases. This is an expected result since we add the effect prediction back into the state vector, effectively cascading the error over multiple steps.

\subsection{Interpreting the Learned Symbols}
\label{subsec:symbols}
In the subfigures of Figure \ref{fig:symbols}, we visualize object positions that activate the symbol (the discrete bottleneck vector) given in the title of the subfigure, together with its activation count. The symbol `0101' is activated only with long blocks, while other symbols respond to short blocks. Symbols `1110' and `1100' are activated for short blocks below the grasped object (as the object position is relative to the grasped object). Likewise, symbols `0111', `1101', and `1111' are primarily active for short blocks above the grasped object. These symbol groundings show that the object type and its relative $z$-axis position are the most significant factors in predicting the effect. On the other hand, no symbol is specialized on the $x$-axis position as it does not bring any additional advantage for the effect prediction.

In the subfigures of Figure \ref{fig:relations}, we visualize the relative positions of objects that activate a relation. We choose to visualize with respect to the relative distance between objects as other factors (e.g., object types, positions) do not seem to be significant on their own. The third relation (R2 in Figure \ref{fig:relations}) is activated only when the objects are aligned within the $y$ axis. This suits the environment since objects are only aligned when they are mostly on top of each other. Furthermore, this alignment information helps the model to differentiate between two different stacks of objects with the same set of objects; objects aligned in the $y$ axis are in the same stack. Other relations seem to be activated for a wide range of relative positions and object types, suggesting that they might not be as significant as the third relation.

\section{CONCLUSION}
\label{sec:conc}
In this paper, we proposed a new method to simultaneously learn object symbols and relations between objects in a single architecture. Namely, \emph{discrete} attention weights are computed from object features to model relations between objects. As these weights are discrete, they can be regarded as relational symbols between objects. Such a feature is desirable because it allows us to model the environment with object symbols and relations between objects, which was not available previously \cite{ahmetoglu2022deepsym,ahmetoglu2022learning}. We showed that the proposed model achieves significantly lower errors than others in predicting the effects of (possibly a sequence of) actions on a varying number of objects, and produce meaningful symbols that allow us to model the relations between objects for settings where the number of objects can vary.

As the next step, we plan to convert the learned symbols into PDDL operators \cite{aeronautiques1998pddl} for domain-agnostic planning with off-the-shelf planners. Rules defined with learned symbols can be generated by a tree learning approach in which the features would be the binding of variable names to the symbol values, and the labels would be unique symbolic effects (unique changes in object symbols and relations). Alternatively, these operators can be learned by partitioning the symbolic dataset as in \cite{chitnis2022learning}. Such a conversion will remove the cascading of errors in action sequence prediction and allow for a fast search in the symbolic space by removing the need to use a neural network.


\section*{ACKNOWLEDGMENT}
This research was supported by TUBITAK (The Scientific and Technological Research
Council of Turkey) ARDEB 1001 program (project number: 120E274). Additional support was given by the Grant-in-Aid for Scientific Research (project no 22H03670), the project JPNP16007 commissioned by the New Energy and Industrial Technology Development Organization (NEDO), and JST, CREST (JPMJCR17A4).

\bibliographystyle{IEEEtran}
\bibliography{IEEEabrv,ref}

\end{document}